\documentclass[lettersize,journal]{IEEEtran}
\usepackage{amsmath,amsfonts}
\usepackage{algorithmic}
\usepackage{algorithm}
\usepackage{array}
\usepackage[caption=false,font=normalsize,labelfont=sf,textfont=sf]{subfig}
\usepackage{textcomp}
\usepackage{stfloats}
\usepackage{url}
\usepackage{verbatim}
\usepackage{graphicx}
\usepackage{cite}
\usepackage{multirow}
\begin{document}

\title{V2X-AHD:Vehicle-to-Everything Cooperation Perception via Asymmetric Heterogenous Distillation Network}

\author{Caizhen He, 
Hai Wang,~\IEEEmembership{Senior Member,~IEEE}
Long Chen,
Tong Luo,
Yingfeng Cai,~\IEEEmembership{Senior Member,~IEEE}

\thanks{This work was supported in part by the National Natural Science Foundation of China (52225212, U20A20333, 52072160), Key Research and Development Program of Jiangsu Province (BE2019010-2, BE2020083-3).}
\thanks{(Corresponding Author: Hai Wang);}
\thanks{C. H. and H. W. are with the School of Automotive and Traffic Engineering of Jiangsu University, Zhenjiang, 212013, China. (e-mail: 2112204007@stmail.ujs.edu.cn, wanghai1019@163.com,) }
\thanks{L. C., and Y.C.are with the Automotive Engineering Research Institute of Jiangsu University, Zhenjiang, 212013, China. (e-mail: chenlong@ujs.edu.cn, caicaixiao0304@126.com)}
\thanks{T. L. is with the School of Automobile and Traffic Engineering of Jiangsu University of Technology, Changzhou, 213001, China. (e-mail: 15951270769@163.com)}
}



\maketitle

\begin{abstract}
Object detection is the central issue of intelligent traffic systems, and recent advancements in single-vehicle lidar-based 3D detection indicate that it can provide accurate position information for intelligent agents to make decisions and plan. Compared with single-vehicle perception, multi-view vehicle-road cooperation perception has fundamental advantages, such as the elimination of blind spots and a broader range of perception, and has become a research hotspot. However, the current perception of cooperation focuses on improving the complexity of fusion while ignoring the fundamental problems caused by the absence of single-view outlines. We propose a multi-view vehicle-road cooperation perception system, vehicle-to-everything cooperative perception (V2X-AHD), in order to enhance the identification capability, particularly for predicting the vehicle's shape. At first, we propose an asymmetric heterogeneous distillation network fed with different training data to improve the accuracy of contour recognition, with multi-view teacher features transferring to single-view student features. While the point cloud data are sparse, we propose Spara Pillar, a spare convolutional-based plug-in feature extraction backbone, to reduce the number of parameters and improve and enhance feature extraction capabilities. Moreover, we leverage the multi-head self-attention (MSA) to fuse the single-view feature, and the lightweight design makes the fusion feature a smooth expression. The results of applying our algorithm to the massive open dataset V2Xset demonstrate that our method achieves the state-of-the-art result. The V2X-AHD can effectively improve the accuracy of 3D object detection and reduce the number of network parameters, according to this study, which serves as a benchmark for cooperative perception. The code for this article is available at https://github.com/feeling0414-lab/V2X-AHD.
\end{abstract}


\section{Introduction}
\IEEEPARstart{E}{nvironmental} perception is a fundamental component of autonomous driving \cite{1}\cite{2}, as it provides information for path planning and motion control \cite{40}\cite{41}\cite{42}. As a critical task of environment perception, object detection \cite{43}\cite{44} has always been a research hotspot in autonomous driving. Light changes, as well as rainy and foggy weather, have a significant impact on vision-based object detection methods, resulting in low-security redundancy. In recent years, with the enhancement of information processing capabilities and the decrease in sensor prices,  lidar \cite{14}\cite{16}\cite{17}\cite{20}\cite{23} has achieved leapfrog development with its superior space construction and anti-weather interference capabilities. Academic and industrial circles are largely in agreement that lidar plays a role in advanced intelligent driving technologies. Despite this, the current application of lidar is limited to the single-vehicle dimension, and single-view perception has many inherent flaws. For instance, in crowded road conditions, it is easy to develop blind spots due to occlusion, whereas in long-distance open road conditions, it will result in sparse perception. More importantly, if a pedestrian suddenly emerges from the blind spot, the self-driving vehicle will have very little time to react, which will be a difficult problem for human drivers as well.

\begin{figure}[!t]
\centering
\includegraphics[width=3.5in]{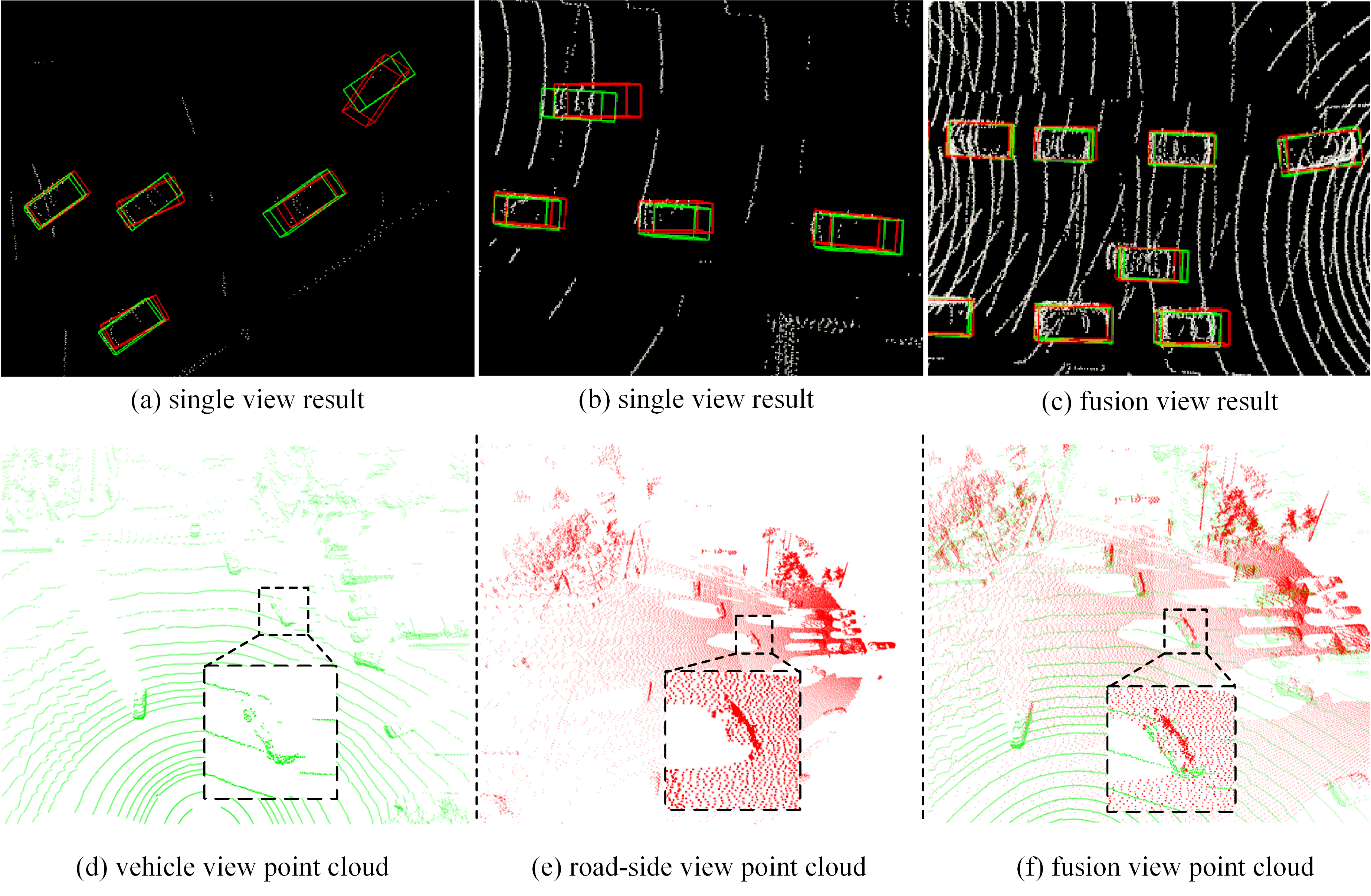}
\caption{Advantages of cooperation perception. The purpose of our model is that a single view of an object has the characteristics of a fusion view. (a) and (b) show the single-view object detection results, respectively. (c) shows the fusion view detection results. (d) and (e) show the single-view vehicle’s point cloud. (f) shows the fusion view vehicle’s point cloud.}
\label{fig_1}
\end{figure}

To address the aforementioned issues with single-view perception, a collaborative perception system incorporating multi-view data has emerged. In congested road conditions, the collaborative sensing system has a global vision. Cross-viewing compensates for the obstruction caused by a single perspective. In contrast, when sensing long-distance road conditions, the collaborative sensing system increases the sensing distance through information relays and solves the issue of sparse perception. It is believed that cooperative sensing can achieve superior perception results compared to single-view vehicles. According to the different transmission content, the existing cooperative sensing strategies can be categorized into three groups based on the transmission content: early fusion\cite{25} of the original point cloud data transmission, intermediate fusion \cite{8}\cite{26}\cite{27}\cite{28} of the transmitted point cloud feature data, and late fusion \cite{29}\cite{30}\cite{39} of the transmission perception results. In the early fusion, the original point cloud data are used for splicing. The perception problem of a single vehicle can be fundamentally resolved by filling the point cloud gap caused by occlusion and expanding the point cloud range. However, due to the enormous amount of data in the original point cloud, the transmission bandwidth limits information transmission, so early fusion cannot be realized. In contrast, the post-fusion strategy transmits each perspective's perception results directly, and the transmission bandwidth requirement is small. However, due to differences in the perception results of the same object from different perspectives, the fusion results are susceptible to the negative results of a single view, resulting in missed and false detections. After the information exchange, the intermediate fusion strategy fuses the extracted intermediate features. It is a compromise strategy that balances perception accuracy and transmission bandwidth, and it has become the predominant research direction of collaborative sensing.

Currently, the intermediate fusion strategy improves perception accuracy by increasing the module's complexity. However, it only improves the fitting capability brought on by the increase in parameters. The characteristics and essence of the fusion perspective have not yet been explored by the aforementioned methods, and increasing the number of model parameters will decrease real-time performance. The accuracy of collaborative perception is dependent on two factors: the extraction of single-view scene features and the fusion of multi-view scene features. Existing intermediate strategy fusion methods emphasize the latter, often ignoring the fundamental role of the former. In our primary research on object detection, we identified two features: First, as shown in Figures 1(a) and (b), the current single-vehicle object detection is more accurate for vehicle position detection but less accurate for vehicle shape detection. Second, as shown in Figure 1(c), the perceptual accuracy of early fusion strategies is typically greater than that of intermediate fusion strategies. The causes for the aforementioned characteristics are also evident. The point cloud of a single vehicle is sparse and incomplete. At the same time, due to the limitation of the viewing angle. As shown in Figures 1(d) and (e), although the algorithm can determine the precise position of the vehicle, it is difficult to determine the car's outline. In Figure 1(f), the vehicle's outline is complete and distinct in the fused viewpoint cloud. The early fusion strategy of direct matching of the original point cloud is the most effective for performance improvement and has interpretability. As shown in Figure 1(c), the roadside radar has more wiring harnesses and denser point clouds, and the collaborative perception of vehicle-road coordination is typically more robust.

At present, the intermediate fusion strategy improves perception accuracy by increasing the complexity of the fusion module. However, it only enhances the fitting ability brought about by the increase in the number of parameters. The above methods have yet to explore the characteristics and essence of the fusion perspective, and increasing the number of parameters of the model will lead to a decrease in real-time performance. The accuracy of collaborative perception depends on two aspects: single-view scene feature extraction and multi-view scene feature fusion. Existing intermediate strategy fusion methods emphasize the latter, often ignoring the fundamental role of the former. In the primary research on object detection, we found two characteristics: First, the existing single-vehicle object detection is more accurate for vehicle position detection, but weak for vehicle shape detection, as shown in Figure 1(a)(b). Second, the perceptual accuracy based on early fusion strategies is usually higher than intermediate fusion strategies, as shown in Fig. 1(c). The reasons for the above characteristics are also apparent. The point cloud of a single vehicle is sparse and incomplete. At the same time, due to the limitation of the viewing angle. Although the algorithm can judge the vehicle's specific position, the car's outline is difficult to determine, as shown in Figure 1(d)(e). In Figure 1(f), the vehicle's outline in the fused viewpoint cloud is clear and complete. The early fusion strategy of direct matching of the original point cloud is the most effective for performance improvement and has interpretability. As shown in Figure 1(c), the roadside radar has more wiring harnesses and denser point clouds, and the collaborative perception of vehicle-road coordination often has a more robust performance.

\begin{figure*}[!t]
\centering
\includegraphics[width=7in]{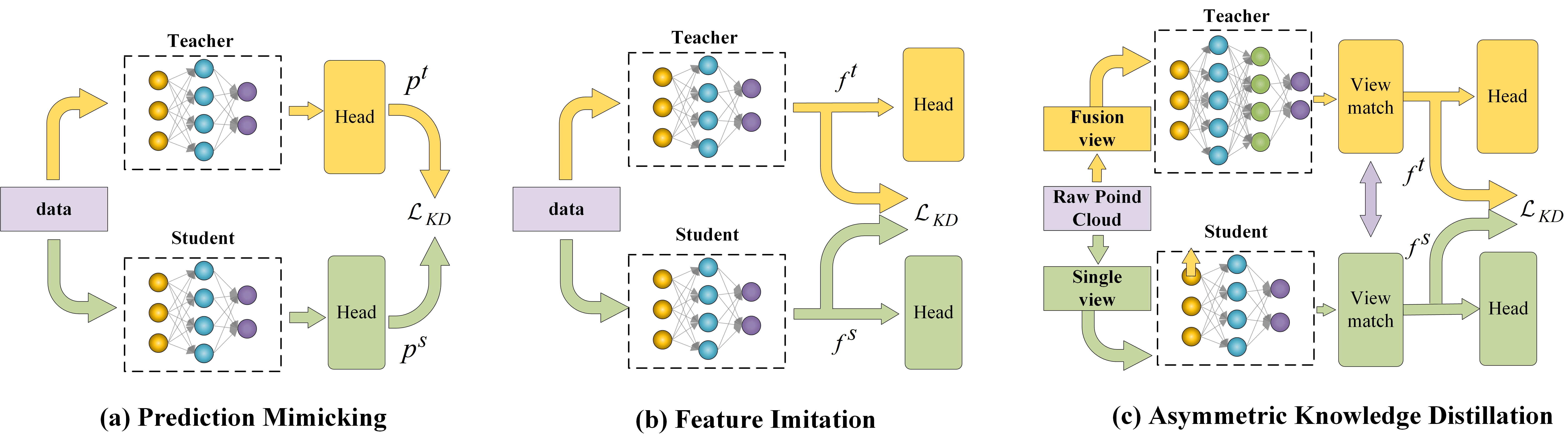}
\caption{Comparison of distillation modes. The yellow and green lines represent the teacher and student network data flow, respectively. (a) and (b) illustrate two kinds of distillation modes, prediction mimicking, and Feature Imitation. (c) is our new mode of asymmetric knowledge distillation, the raw point cloud is divided into a fusion view and a single view.}
\label{fig_2}
\end{figure*}

This paper proposes an algorithm for vehicle-to-everything cooperative perception V2X-AHD based on an asymmetric heterogeneous distillation network. Among them, asymmetry refers to the difference between the parameters of the teacher network and the student network, while heterogeneity refers to the heterogeneity of point cloud data generated by other sensor models and different detection perspectives due to vehicle-side data and roadside data. The main contributions of this paper are as follows:

1. Propose an asymmetric structure for knowledge distillation \cite{3}\cite{4}\cite{5}. Integrate the strategic benefits of early and intermediate integration. During training, the teacher and student networks are separately trained using fused point cloud and single view data. When using single-view data to extract features, the student network has the same fusion point cloud features as the teacher network and can automatically complete the incomplete vehicle outline features. The asymmetric structure is adopted for the difference of cross-view point cloud data of vehicles and roads, which improves the accuracy of point cloud feature extraction with different structures.

2. Feature extraction module Spare Pillar. Due to the sparse nature of point clouds, we convert them to 2D pseudo-images and design a Spare Pillar feature extraction module based on sparse convolution \cite{7}. Compared with the conventional dense convolution method, it has superior feature encoding capabilities while significantly reducing computation time. The step-by-step descending structure of the bottleneck module facilitates multi-scale feature fusion and helps decouple point cloud features of different sizes.

3. Multi-head self-attention (MSA) feature fusion module. Existing methods hinder the practical expression of single-vehicle features due to the use of complex fusion modules; therefore, we employ a lightweight MSA mechanism to reduce computation while realizing the practical expression of single-vehicle features.
The remainder of the paper is organized as follows. The literature related to this work is briefly introduced in related work. The Methodology section describes the V2X-AHD algorithm structure proposed in this paper. The Experiments section demonstrates the results of the experiments of the proposed algorithm on the V2XSet\cite{8} dataset, while the Conclusion section provides a summary.

\section{related works}
\subsection{Knowledge Distillation}
Knowledge distillation is also known as the teacher-student model \cite{3}\cite{4}\cite{5}. Geoffrey Hinton \cite{4} noted that the smooth distribution result predicted by the teacher network makes it easier for students to learn features than the Dirac result. The knowledge distillation network trains the teacher and student networks with different parameters. The network of teachers with a large parameter scale can be compressed into a smaller network of students. At the same time, the teacher network and the student network can input different data for training. Currently, this method is widely employed in the fields of semantic segmentation \cite{10}, point cloud target detection \cite{11}, and object re-identification \cite{12}. As shown in Figures 2(a) and 2(b), the distillation network can be divided into Feature Imitation and Prediction Mimicking, depending on the transfer feature content. DiscoNet \cite{13} utilizes a symmetric distillation network and a predictive simulation structure to transform multi-view point cloud features into a single-view result. However, DiscoNet's Prediction Mimicking has the problem of hindering the detection of performance diversity. In contrast, the purpose of feature simulation is to improve the consistency of the teacher-student model in terms of latent features. As shown in Figure 2(c), we employ an innovative asymmetric knowledge distillation structure to train the point cloud feature extraction. Compared with the student network, the teacher network has a more significant number of parameters and uses multi-view fusion point clouds as data input. The parameters of the student network are small, and single-view point cloud data are used to extract the network's features. During the stage of training, the fused-view features of the teacher network are transferred to the single-view features of the student network. During the test phase, even without the guidance of the teacher network, the student network retains the feature expression of the fusion point cloud beneath the single-view data.

\subsection{3D point cloud object detection}

Currently, the 3D object detection algorithms based on point cloud follow the "encoding-decoding-detection head" detection architecture and adopt the grid-based form. According to the type of feature extraction convolutional network, it can be divided into two categories: the first category, based on the 3D voxel grid method \cite{14}\cite{15}\cite{16}\cite{17}\cite{18}, uses a 3D convolutional network to extract features; the second category, based on 2D Pillar grids, is the method \cite{19}\cite{20}\cite{21}\cite{22}, uses 2D convolutional networks to extract features. 
VoxelNet[18], the pioneering work of the voxel grid method, divides the point cloud space into 3D voxels and encodes point cloud features using 3D convolution. The Second \cite{16} method investigates the problem of empty voxels in the VoxelNet method's long-distance area and employs a sparse 3D convolution method to improve the detection accuracy. The above algorithms that use 3D convolutional networks to extract features have high accuracy. However, due to the high computational complexity of the 3D convolution method, even the enhanced sparse method still has a substantial computational overhead. 
To address the aforementioned issues, a class of techniques employing 2D convolutions has gradually emerged, converting 3D point clouds into 2D pseudo-image features and then employing 2D convolutions with reduced operational complexity. One representative algorithm is the PointPillar \cite{19} pillar feature extractor based on PointNet \cite{23}. HVNet \cite{21} combines columnar features of varying scales to enhance operational precision and inference speed.
Currently, algorithms based on 2D pseudo-graph features prioritize the projection of complex pillar features and multi-scale aggregation. However, compared to images, the information density of point cloud grids is significantly smaller than that of image grids, and the features are sparse. Using dense convolution methods for point cloud features wastes computing resources. We use sparse 2D convolution as the core and propose the Spare Pillar Plug and Play feature extraction network. Sparse convolution can reduce computation time significantly. At the same time, the detection accuracy improves due to the more concentrated feature extraction.

\begin{figure*}[!t]
\centering
\includegraphics[width=7in]{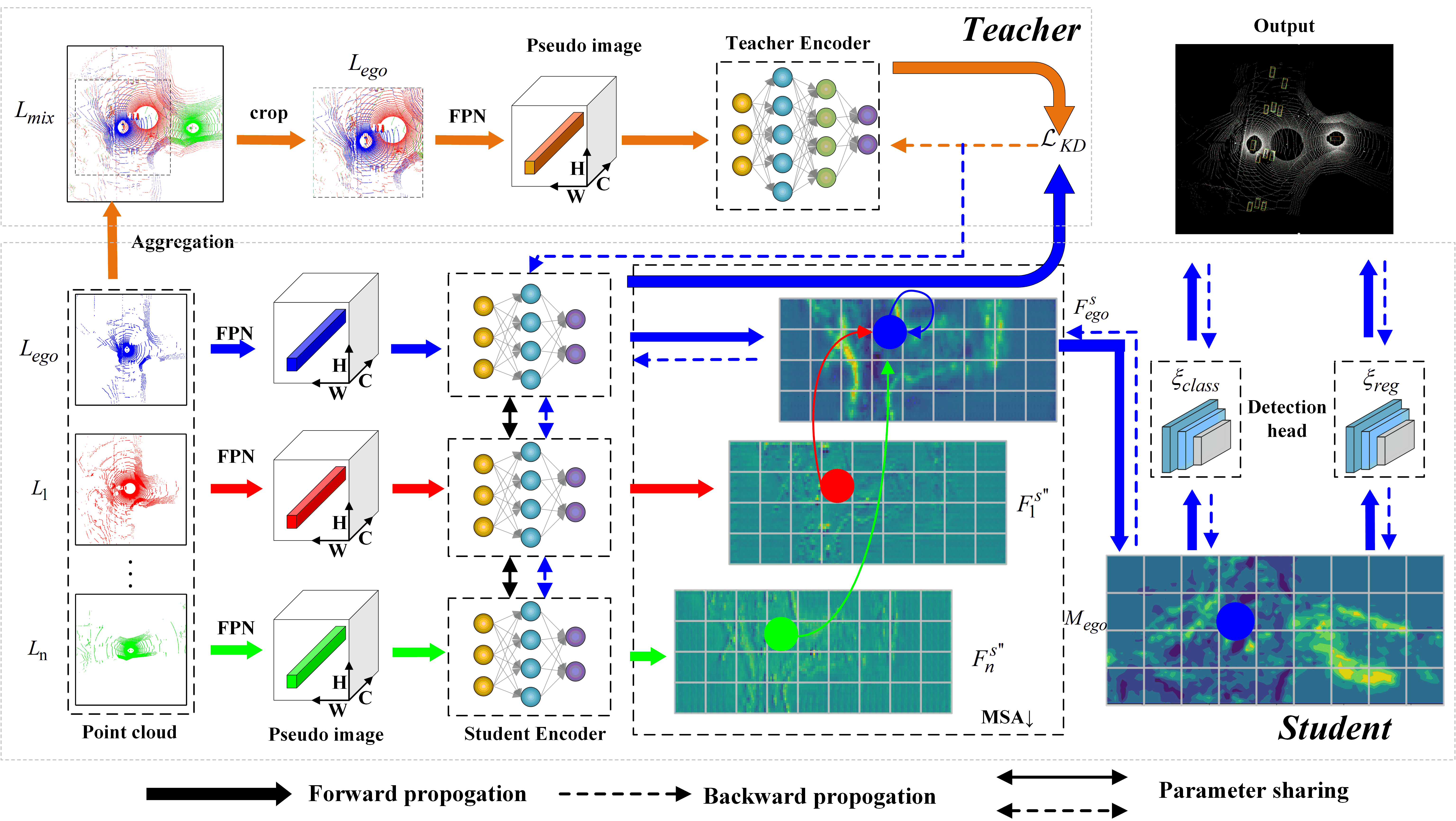}
\caption{Pipeline of our proposed V2X-AHD. It consists of two networks: a  teacher network and a student network. The single arrow solid line and dotted line represent the forward propagation and backward propagation process, respectively. The details of each individual component are illustrated in Sec.3.}
\label{fig_3}
\end{figure*}

\subsection{ Vehicle-to-Everything Cooperation Perception}
Vehicle-to-Everything collaborative perception is the process of obtaining fusion perception results by receiving the perception data of neighboring agents. Due to its finer point cloud data and more stable own position coordinates, roadside lidar is more conducive to enhancing detection accuracy than onboard lidar. However, because access to roadside equipment will result in differences in point cloud data perspectives, a heterogeneous fusion module is required to fuse them.
Currently, the mainstream algorithm of collaborative sensing can be divided into three categories based on the various transmission data types: the early fusion of the original point cloud data \cite{25}, the intermediate fusion of the transmission point cloud features \cite{8}\cite{26}\cite{27}\cite{28}, and the transmission perception. As a result of late fusion \cite{29}\cite{30}\cite{39}, the required bandwidth for data transmission decreases for each of the three methods. Cooper[25] proposed the early fusion method based on the original point cloud data level. He obtained the perception result of the complete perspective by aggregating the surrounding point cloud data. This method can fundamentally solve the problem of a single perspective, but transmitting raw point cloud data requires a substantial amount of communication bandwidth and high latency. In the late fusion algorithm, Andreas\cite{29} proposed a high-dimensional sensor fusion structure that uses temporal and spatial sequences as input data to predict the position, action, and angle of surrounding agents in three dimensions. Rawashdeh \cite{30} utilized machine learning techniques to transmit three-dimensional information, including the position and size of the tracking object's center point, in order to accurately predict surrounding objects. However, in late fusion, although the required data transmission bandwidth is small, a significant amount of effective scene information will be lost, resulting in excessive dependence on the results of a single agent, and an effective error correction mechanism cannot be created when individual sensors make errors. 
The intermediate fusion strategy can reduce the transmission bandwidth requirements to reduce the loss of scene information, which has attracted a large number of researchers to conduct research in this area. F-cooper \cite{26} employs voxel characteristics for information transfer and the Maxout \cite{31} method for voxel fusion at the junction. The OPV2V \cite{28} algorithm leverages the self-attention mechanism \cite{32} to learn the interaction between compressed features at the same spatial position. The V2Vnet \cite{27} algorithm fuses passed compressed features using a graph convolutional neural network GNN \cite{33} network. V2X-ViT \cite{8} considers the issue of space-time misalignment caused by real-world communication delays and employs self-attention mechanisms and multi-scale windows to fuse features from various perspectives. The MSA module minimizes the parameters' size while maintaining the perception's accuracy.

The following sections will focus on the structural details of each part of the model V2X-AHD proposed in this paper.

\section{Methodology}

In this paper, we propose a Vehicle-to-Everything cooperation perception system V2X-AHD based on 3D point cloud data to perceive the surrounding environment of a vehicle. The primary sources of design inspiration are the high precision of the early fusion strategy and the benefits of the knowledge distillation network for joint training of different input stream data. The advantage of the knowledge distillation system is its ability to compress the knowledge gained from the large model into the small model by means of "knowledge distillation." At the same time, the distillation model can be used for zero-shot or few-shot learning, allowing the teacher and student networks to input distinct training data. The point cloud of Vehicle-to-Everything cooperation perception is unique, and it is simple to obtain multi-view point cloud feature training offline. In contrast, in the online test scenario, the agent can only accept a single perspective point cloud to extract and transmit features, which is perfectly compatible with the distillation network.

During the training stage, the teacher model uses multi-view fusion data, while the student model uses single-view data. Both of them use the Spare Pillar to extract the points cloud feature. Through the distillation method of feature simulation, the teacher model transfers the multi-view feature extraction paradigm to the single-view student model. During the test stage, although the student model is trained with single-view data, the detection performance can approach that of the teacher model using multi-view data, thereby fundamentally solving the problem of single-view data feature extraction.

\begin{figure*}[!t]
\centering
\includegraphics[width=7in]{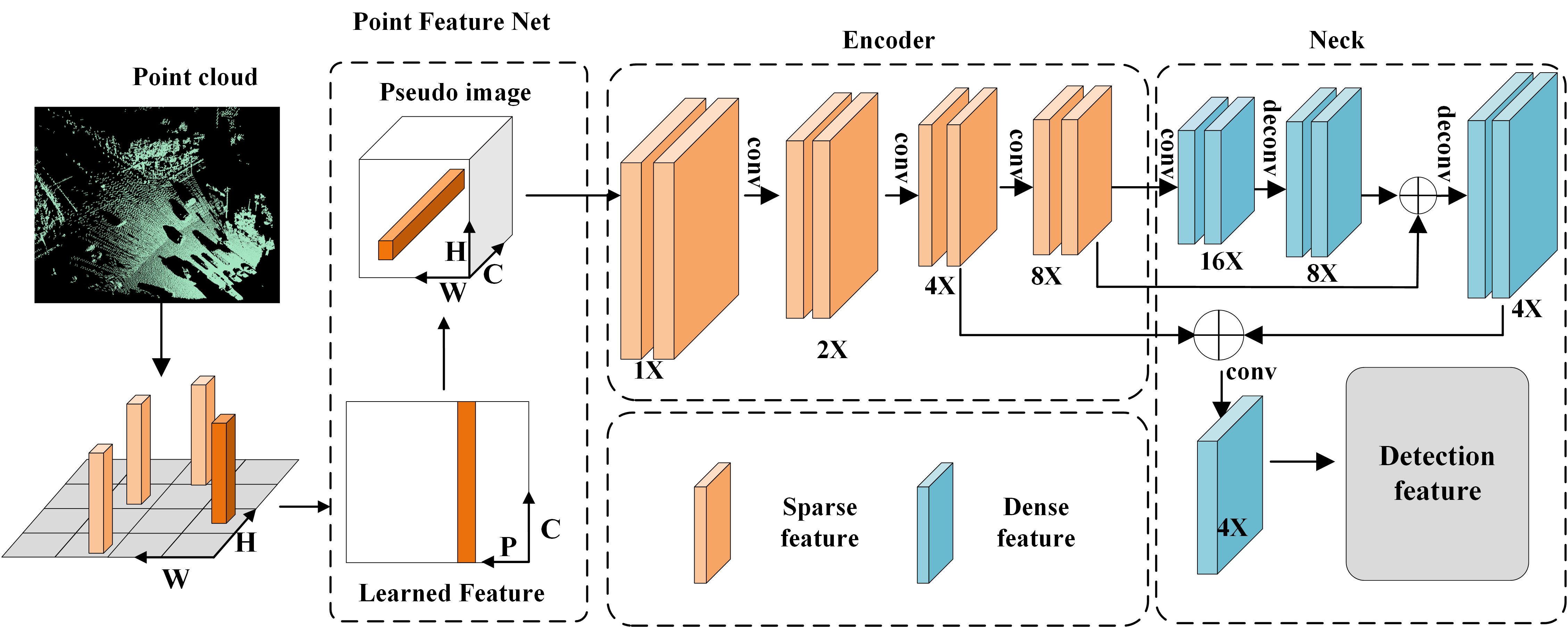}
\caption{Feature extraction network Spare Pillar structure. The structure includes three parts: a point feature net, an encoder, and a neck. Orange and blue modules represent sparse convolution and dense convolution, respectively}
\label{fig_4}
\end{figure*}

\subsection{Student network}

Each agent \textbf{\textit{i$\in$\{1,...,N\}}} within the communication distance, agent categories \textbf{\textit{c$_i$$\in$\{V,I\}}}, \textbf{\textit{V}} and \textbf{\textit{I}} represent vehicles and roadside equipment, respectively. We assume that the data transmission is synchronized, which means that each agent \textbf{\textit{i}} position is \textbf{\textit{P$_i$}}, and the lidar data are \textbf{\textit{L$_i$}}. Assuming that agent \textbf{\textit{ego}}is selected as the central agent, and \textbf{\textit{ego}} accepts the positions from surrounding agents. Central agent \textbf{\textit{ego}} can receive the original point cloud or feature from surrounding agents through coordinate system transformation.

\subsubsection{Feature extraction}

The point cloud feature extraction algorithm uses the 2D columnar grid feature extraction algorithm Spare Pillar, as shown in Figure 4. Using the "Encoder-Neck" architecture under the Bird's Eye View (BEV) while retaining the traditional 2D columnar grid algorithm, it employs a sparse convolution method better suited for point cloud features to further reduce inference delay and improve feature extraction capabilities. The overall structure consists of three components: a 2D pseudo-image generation module, encoder, and bottleneck block Neck. Point Feature Net converts the point cloud into a stacked Pillar tensor, which is then projected into a 2D pseudo-image with a scale of \textbf{\textit{H$\times$W$\times$C}}, where \textbf{\textit{H}} and \textbf{\textit{W}} represent the height and width of the pseudo-image canvas, respectively, and A indicates the number of channels of the pseudo-image. The 2D pseudo-image is then fed into the encoder. The Encoder adopts the VGGNet \cite{34} architecture. The objective is to extract sparse columnar features of varying depths from the projected sparse 2D pillar and to feed all sparse pillar features of varying scales into the bottleneck block. Since feature compression has been performed, sparse pillar features can be fused using standard dense 2D convolutions. The specific procedure is illustrated by formula \eqref{equation1}:

\begin{equation}
\label{equation1}
F_i^s = \phi_s(L_i),F_i^s\in\mathbb R^{\bar{H}\times\bar{W}\times\bar{C}}
\end{equation}

where $\phi_s(\bullet)$ represents the student feature extraction network, and $\mathbb R^{\bar{H}\times\bar{W}\times\bar{C}}$ indicates the scale of the feature space after convolution. SparePillar employs sparse convolution to reduce computational complexity during the stage of feature extraction. In the feature upsampling procedure, dense convolution is used to integrate high-level abstract semantics and low-fine-grained spatial features to improve the accuracy of large objects. We use different feature extraction structures for the teacher and student models. The experimental section will demonstrate distinct structural differences.

\subsubsection{Compression and decompression}

To further reduce the required bandwidth for data transmission, each agent needs to perform compression before data transmission. We use 1$\times$1 convolutional layer to compress the features in the channel direction, as shown in formula \eqref{equation2}:

\begin{equation}
\label{equation2}
F_i^{s'} =Enc_{com}(F_i^s), F_i^{s'}\in\mathbb R^{\bar{H}\times\bar{W}\times\bar{\bar{C}}},\bar{\bar{C}}\ll\bar{C}
\end{equation}
\begin{equation}
\label{equation3}
Data_i\leftarrow(F_i^{s'},P_i)
\end{equation}

where $Enc_{Com}(\bullet)$ represents the compression function, formula \eqref{equation3} represents the information transmission data packet, $Data_i$ transmits the compressed feature  $F_i^{s'}$ and its position  $P_{i}$, and decompresses after other agents receive the compressed feature $F_i^{s'}$; the specific process is shown in formula \eqref{equation4}:

\begin{equation}
\label{equation4}
F_i^{s''}=Dec_{com}(F_i^{s'}),F_i^{s''}\in\mathbb R^{\bar{H}\times\bar{W}\times\bar{C}}
\end{equation}

where $Dec_{Com}(\bullet)$ represents the decompression function corresponding to the compression process, and the decoded feature space becomes  $\mathbb R^{\bar{H}\times\bar{W}\times\bar{C}}$. The decompressed feature $ F_i^{s"}$ will be transmitted to the feature fusion part.

\subsubsection{Feature fusion}

The MSA mechanism model \cite{32} fuses the decompressed features. The specific procedure is shown in Figure 5. The feature vectors at the same position on the feature map correspond to particular points in the original point cloud data. Spatial correlation is destroyed by tiling the feature map and calculating the weighted sum of each feature. However, although the complex fusion structure objectively increases the network's depth and thus improves its ability to fit, this operation will hinder the accurate expression of individual features. We use an MSA mechanism with a few parameters to more accurately capture the features it represents and generate a fusion feature map. The specific procedure is shown in formulas \eqref{equation5} and \eqref{equation6}:

\begin{equation}
\label{equation5}
m_i = M(P_{ego},P_i)
\end{equation}

\begin{equation}
\begin{aligned}
\label{equation6}
M_{ego} = MSA(F_{ego}^s,(F_1^{s''},m_1),\\(F_2^{s''},m_2),\dots ,(F_n^{s''},m_n))
\end{aligned}
\end{equation}

The formula \eqref{equation5} indicates that after receiving the data $Data_i$, $M(\bullet)$ means to use its position information $P_{ego}$ and $P_i$ to calculate the position transfer matrix $m_i$. After that, we send matrix $m_i$ and the decompressed feature $F_1^{s"}$ to the fusion network $MSA(\bullet)$, as shown in the formula \eqref{equation6}. $MSA(F_{ego}^s$ is the feature of its point cloud, and  $M_{ego}$ represents the fusion feature. Finally, the fusion features are sent to the detection head.

The fusion network structure is shown in Figure 5. After the features have been connected, the MSA mechanism is used to extract features, where A represents the number of multi-head attention heads. Formulas \eqref{equation7}, \eqref{equation8}, and \eqref{equation9} represent the specific execution steps of the multi-head attention mechanism, while formula \eqref{equation7} represents the connection between the multi-head attention feature results; formula \eqref{equation8} represents the generation of a single attention head, where $W_i^Q$, $W_i^K$, and $W_i^V$ represents the relationship matrix of $Q$,$K$,$V$, $K$, and $V$ values between single attention head $Attention(Q',K',V')$ and multi-head attention $Multihead(Q,K,V)$, respectively. Formula \eqref{equation9} represents the operation process of single-head self-attention, where   $dK'$ indicates the feature dimension.

The fusion network structure is shown in Figure 5. After the features are connected, the multi-head self-attention mechanism is used for feature extraction, where the parameter A represents the number of multi-head attention heads. Formulas \eqref{equation7}, \eqref{equation8}, and \eqref{equation9} represent the specific execution steps of the multi-head attention mechanism, and formula \eqref{equation7} represents the connection of multi-head attention feature results; formula \eqref{equation8} represents the generation of a single attention head, where $W_i^Q$, $W_i^K$, $W_i^V$represents the relationship matrix of $Q$,$K$,$V$values between single attention head $Attention(Q',K',V')$ and multi-head attention $Multihead(Q,K,V)$ , respectively. formula \eqref{equation9} represents the operation process of single-head self-attention, where $dK'$ means the feature dimension.

\begin{equation}
\begin{split}
\label{equation7}
Multihead(Q,K,V)=Linear(Concat\\(head_1,head_2,\dots ,head_h))
\end{split}
\end{equation}

\begin{equation}
\label{equation8}
head_i=Attention(QW_i^Q,KW_i^K,VW_i^V)
\end{equation}

\begin{equation}
\label{equation9}
Attention(Q',K',V')=softmax(\frac{Q'K'^T}{\sqrt{dK'}})V'
\end{equation}

\begin{figure*}[!t]
\centering
\includegraphics[width=7in]{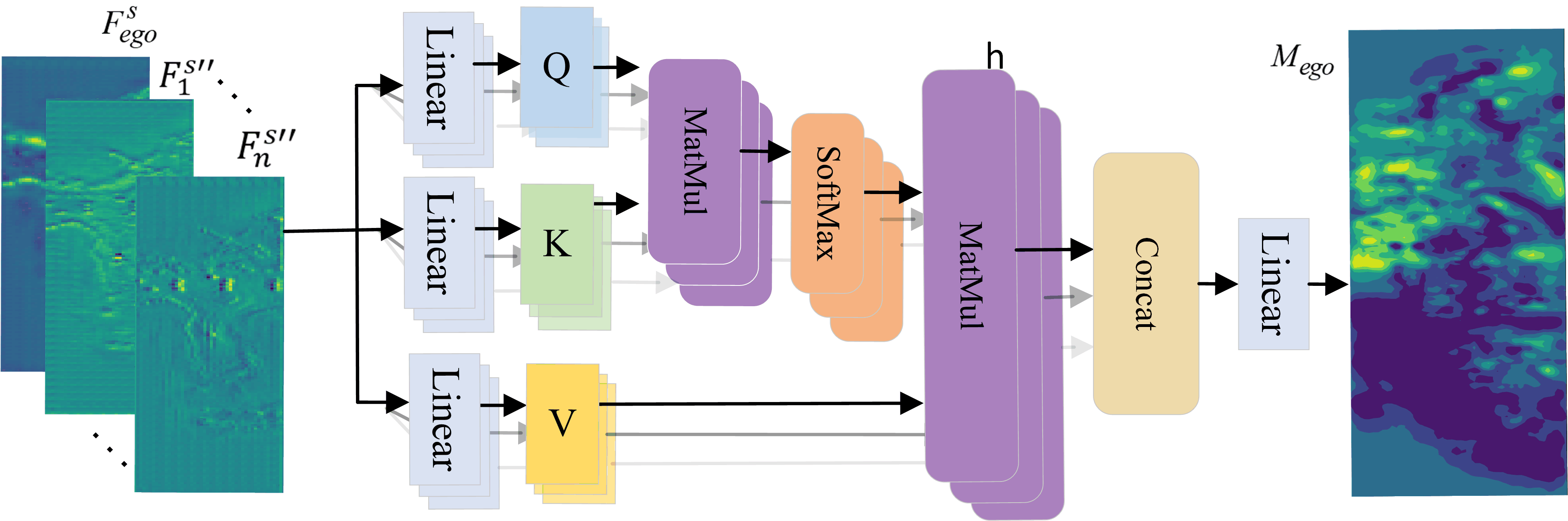}
\caption{Feature fuse module multi-head attention (MSA) mechanism.}
\label{fig_5}
\end{figure*}

\subsubsection{Detection Head}

After obtaining the final fusion feature $M_{ego}$, we use two 1$\times$1 convolutional layers to generate classification and regression prediction results, respectively, and form a prediction box, as shown in formulas \eqref{equation10} and \eqref{equation11}:

\begin{equation}
\label{equation10}
Y_{class} = {\xi}_{class}(M_{ego})
\end{equation}

\begin{equation}
\label{equation11}
Y_{reg} = {\xi}_{reg}(M_{ego})
\end{equation}

where ${\xi}_{class}(\bullet)$  represents the classification layer, and  $Y_{class}$ outputs a score, which is used to indicate whether the preselected box is an object or a background. ${\xi}_{reg}(\bullet)$ represents the regression layer, and the output of $Y_{reg}$ is seven dimensions $(x,y,z,w,l,h,\theta)$ , where $x$, $y$ , and  $z$ represent the position of the prediction frame, $w$, $l$, and $h$ represent the size of the prediction frame, and  $\theta$ represents the heading angle of the prediction box.

\subsection{Teacher network}
\subsubsection{Multi-view data fusion}
The input data of the teacher network are fused perspective data, which needs to be fused before inputting into the model. The process of fused point cloud is as follows:

\begin{equation}
\begin{split}
\label{equation12}
L_{mix} = A((L_{ego},P_{ego}),\\(L_1,P_1),(L_2,P_2),\dots,(L_N,P_N))
\end{split}
\end{equation}

where $L_{mix}$ represents the fused point cloud, $N$ indicates the number of agents within the transmission range, and $A(\bullet)$ represents the aggregation process of the surrounding point cloud data. To ensure that the input data of the teacher network of the fusion view and the student network of a single view are aligned, the coordinates are transformed into the coordinates centered on agent $ego$ after the fusion point cloud is cropped.

\subsubsection{Feature extraction}

After collecting the multi-view data, we feed it into the network for feature extraction. The process of feature extraction for the teacher network is similar to that for the student network, but the input data differ, as shown in formula \eqref{equation13}:

\begin{equation}
\label{equation13}
F_i^t = \phi_t(L_{mix}),F_i^t\in\mathbb R^{\bar{H}\times\bar{W}\times\bar{C}}
\end{equation}

$\phi_t(\bullet)$ represents the feature extraction network of the teacher's point cloud,  $F_{ego}^t$ indicates the feature extracted from the fusion point cloud centered on agent $ego$, and $R^{\bar{H}\times\bar{W}\times\bar{C}}$ represents the same feature space as the student network.

\subsubsection{Knowledge distillation loss function}
During the training process, the teacher network is more straightforward to converge than the student network training; therefore, the teacher and student networks can be jointly trained with random initialization parameters. In this paper, we jointly train the model using object detection loss and distillation loss. The loss ${\mathcal L}_{total}$ that needs to be minimized is shown in formula \eqref{equation14}:

\begin{equation}
\label{equation14}
{\mathcal L}_{total} = {\lambda}_{det}{\mathcal L}_{det}+{\lambda}_{KD}{\mathcal L}_{KD}
\end{equation}

where hyperparameters ${\lambda}_{det}$ and ${\lambda}_{KD}$ control the weights of object detection loss ${\mathcal L}_{det}$ and knowledge distillation loss ${\mathcal L}_{KD}$, respectively. The target detection loss is shown in formula \eqref{equation15}, including classification loss ${\mathcal L}_{class}(\bullet)$ and regression loss ${\mathcal L}_{reg}(\bullet)$. Classification loss ${\mathcal L}_{class}(\bullet)$ use focal loss[38] to calculate classification $Y_{class}(\bullet)$ and label classification value $\hat{Y}_{class}$, which is used to judge whether the object in the detection frame is background or target. Regression loss ${\mathcal L}_{reg}(\bullet)$ uses ${\ell}_1$ smooth loss to calculate regression value $Y_{reg}$ and label regression value $\hat{Y}_{reg}$ , which are used to judge the detection frame's position, size, and heading angle.

\begin{equation}
\label{equation15}
{\mathcal L}_{det} = {\mathcal L}_{reg}(Y_{reg},\hat{Y}_{reg})+{\mathcal L}_{class}(Y_{class},\hat{Y}_{class})
\end{equation}

The distillation loss ${\mathcal L}_{KD}(F_{ego}^t,F_{ego}^s)$ is shown in formula \eqref{equation16}, where $KL((p(x)||(q(x))$ represents the Kullback-Leibler(KL) divergence, which is used to describe the difference between the distributions $p(x)$ and $q(x)$. $V(\bullet)$ indicates retaining the feature channel $\hat{c}$ and generating a one-dimensional feature (vector) from a feature of $\hat{H}\times\hat{W}$ size. The $\tau(\bullet)$ function represents the softmax operation process with the distillation temperature $T$ according to the number of channels $\hat{c}$, as shown in formula \eqref{equation17}.

\begin{equation}
\label{equation16}
{\mathcal L}_{KD}(F_{ego}^t,F_{ego}^s) = KL(\tau(V(F_{ego}^t))||\tau(V(F_{ego}^s)))
\end{equation}

\begin{equation}
\label{equation17}
\tau(\bullet) \leftarrow \frac{exp(z_i/T)}{\sum_jexp(z_j/T)}
\end{equation}

where $z_i$ represents the $i$-th channel's characteristics, and the hyperparameter $T$ represents the distillation temperature.

This section describes the network structure and direction of data flow for each network component. Experiments will be conducted in the next section to confirm the effectiveness of the structure described above.

\section{Experiments}
\subsection{Dataset}
To evaluate the performance of the proposed V2X-AHD algorithm, all experiments use the open large-scale vehicle-road collaboration dataset V2XSet \cite{8} . CARLA \cite{35} and OpenCDA \cite{36} generate V2XSet to simulate natural scenes. The data set consists of 11447 frames, and since each frame contains multiple agents, the total number of samples is 33081. Regarding structure, the training, verification, and test sets are 6694/1820/2833 frames, respectively.

\begin{table}
\renewcommand\arraystretch{1.8}
\tabcolsep=0.1cm
\caption{Detailed architectural specifications for V2X-AHD.}
\label{tab1}
\centering
\begin{tabular}{ c c c }
\hline
name & outputsize& Framework\\

\hline
Teacher-Conv1&[32,192,704]&[SpConv3$\times$3,32$\to$32,stride=1,BN,ReLU]$\times$2\\
Teacher-Conv2&[64,96,352]&[SpConv3$\times$3,32$\to$64,stride=1,BN,ReLU]\\
&&[SpConv 3$\times$3,32$\to$64,stride=1,BN,ReLU]$\times$2\\
Teacher-Conv3&[128,48,176]&[SpConv3$\times$3,64$\to$128,stride=1,BN,ReLU]\\
&&[SpConv 3$\times$3,128$\to$128,stride=1,BN,ReLU]$\times$2\\
Teacher-Conv4&[256,24,88]&[SpConv3$\times$3,128$\to$256,stride=1,BN,ReLU]\\
&&[SpConv 3$\times$3,256$\to$256,stride=1,BN,ReLU]$\times$2\\
Teacher-Conv5&[256,12,44]&[Conv3$\times$3,256$\to$256,stride=1,BN,ReLU]\\
&&[Conv 3$\times$3,256$\to$256,stride=1,BN,ReLU]$\times$2\\
\hline
Student-Conv1&[32,192,704]&[SpConv3$\times$3,32$\to$32,stride=1,BN,ReLU]$\times$2\\
Student-Conv2&[64,96,352]&[SpConv3$\times$3,32$\to$64,stride=1,BN,ReLU]\\
&&[SpConv 3$\times$3,32$\to$64,stride=1,BN,ReLU]\\
Student-Conv3&[128,48,176]&[SpConv3$\times$3,64$\to$128,stride=1,BN,ReLU]\\
&&[SpConv 3$\times$3,128$\to$128,stride=1,BN,ReLU]\\
Student-Conv4&[256,24,88]&[SpConv3$\times$3,128$\to$256,stride=1,BN,ReLU]\\
&&[SpConv 3$\times$3,256$\to$256,stride=1,BN,ReLU]\\
Student-Conv5&[256,12,44]&[Conv3$\times$3,256$\to$256,stride=1,BN,ReLU]\\
&&[Conv3$\times$3,256$\to$256,stride=1,BN,ReLU]\\
\hline
Neck-DeConv1&[128,24,88]&[ConvT2$\times$2,256$\to$128,stride=2,BN,ReLU]\\
&&[Conv3$\times$3,128$\to$128,stride=1,BN,ReLU]\\
Neck-DeConv2&[384,48,176]&[ConvT2$\times$2,256$\to$256,stride=2,BN,ReLU]\\
&&[Conv3$\times$3,256$\to$384,stride=1,BN,ReLU]\\
\hline
MSA&[256,48,176]&[Linear,384$\to256$]\\
&&[Multi head:3,256$\to256$]\\
&&[Linear,256$\to256$]\\
\hline
Detect heads&[16,48,176]&${\xi}_{class} $[Conv1$\times$1,256$\to$2,stride 1]\\
&&${\xi}_{reg} $[Conv1$\times$1,256$\to$14,stride 1]\\
\hline

\end{tabular}
\end{table}

\subsection{Settings}

As an evaluation metric, we employ the general 3D object detection standard: the average precision results when the Intersection-over-Union (IoU) threshold is 0.5 and 0.7. In the experiment, the communication distance between any two agents is 70 m, agents beyond this distance are ignored, and the compression rate of transmitted data is uniformly set to 32. The x- and y-axis measurement ranges are [-140.8, 140.8] and [-38.4, 38.4], respectively, while the z-axis measurement range is [-1,3]. A single voxel's length, width, and height are set to 0.4, 0.4, and 4, respectively. During the training process, all comparison methods employ the Adam \cite{37} optimizer for 60 training rounds; the initial learning rate is set to 0.001, and it is decreased to the original value of 0.1 every 20 training rounds. All models are trained on a device equipped with an NVIDIA GeForce RTX 3090 graphics card and an AMD 5900x CPU. \ref{tab1} displays the network and parameter information for the distillation network.

\subsection{Comparison methods}
Early Fusion, which aggregates the original point clouds of surrounding agents; Late Fusion, which obtains the perception results of all surrounding agents and employs the non-maximal suppression method to produce the final result; This experiment compares intermediate fusion strategies by contrasting the most recent five algorithms: F-cooper, V2VNet, DiscoNet, OPV2V, and V2X-ViT. For a fair comparison, the same method of feature extraction is used during evaluation. As the backbone for testing, they are based on the PointPillar \cite{19} and SparePillar methods proposed in this paper.

\begin{table}
\renewcommand\arraystretch{1.8}
\tabcolsep=0.17cm
\caption{Detection performance comparison on V2Xset.We show average precision (AP) at IoU=0.5,0.7 on Point Pillar and Spare Pillar Backbone, respectively.}
\label{tab2}
\begin{tabular}{ c c c c c c} \hline
\multicolumn{1}{c}{\multirow{2}{*}{Methods}} &\multicolumn{1}{c}{\multirow{2}{*}{Source}}&\multicolumn{2}{c}{Point Pillar AP $\uparrow$}&\multicolumn{2}{c}{Spare Pillar AP $\uparrow$}\\ \cline{3-6}
&& IoU=0.5 & IoU=0.7 & IoU=0.5 & IoU=0.7 \\ \hline
Early Fusion& &0.877 & 0.791	& 0.924 & 0.836\\
Late Fusion& &0.727 & 0.620	& 0.755 & 0.658\\
No Fusion& &0.606 & 0.402	& 0.651 & 0.573\\\hline
F-Cooper\cite{26}&2019 SEC&0.840 & 0.680	& 0.872 & 0.737\\
V2VNet\cite{27} &2020 ECCV&0.845 & 0.677	& 0.832 & 0.673\\
DiscoNet\cite{13} &2021 NeurIPS&0.844 & 0.695	& 0.857 & 0.726\\
OPV2V\cite{28} &2022 ICRA&0.807 & 0.664	& 0.864 & 0.754\\
V2X-ViT\cite{8} &2022 ECCV&\textbf{0.882} & 0.710	& 0.854 & 0.742\\\hline
V2X-AHD &OURS&0.855 &\textbf{ 0.724}	& \textbf{0.877} &\textbf{ 0.770}\\\hline
\end{tabular}
\end{table}

\subsection{Quantitative analysis}
In this section, we present the results of our V2X-AHD method and compare it to the literature.
\subsubsection{Main methods results}
The experimental results of existing 3D object detection algorithms are compared in \ref{tab2}. At the same time, the Spare Pillar module proposed in this paper is a plug-and-play module. The test results indicate that the point pillar and the spare pillar should serve as the network's backbone. The experimental results demonstrate that the intermediate strategy fusion methods have better results than the no-fusion and late-fusion algorithms. The early fusion strategy algorithm can receive complete original point cloud data, and the test results are better than the intermediate fusion strategy algorithm. Under IoU=0.7, the proposed V2X-AHD algorithm achieves the best performance. Compared with the optimal algorithm V2X-ViT algorithm, using the Point Pillar and Spare Pillar backbone networks, the accuracy is increased by 1.6\% and 2.2\%, respectively. The perception accuracy of the Spare Pillar module is superior to that of the Point Pillar module. When IoU=0.7, the results of Fcooper, DiscoNet, OPV2V, and V2X-ViT methods increased by 8.4\%, 4.5\%, 13.1\%, and 4.5\% respectively.

Table 2 shows the comparative experimental results of existing 3D object detection algorithms. At the same time, the Spare Pillar module proposed in this paper is a plug-and-play module. The test results show using the point pillar and the spare pillar as the backbone network. The experimental results show that the intermediate strategy fusion methods have better results than the no-fusion and late-fusion algorithms. The early fusion strategy algorithm can receive complete original point cloud data, and the test results are better than the intermediate fusion strategy algorithm. Under IoU=0.7, the V2X-AHD algorithm proposed in this paper achieves the best results. Compared with the optimal algorithm V2X-ViT algorithm, using the Point Pillar and Spare Pillar backbone networks, the accuracy is increased by 1.6\% and 2.2\%, respectively. The Spare Pillar module has higher perception accuracy than the Point Pillar module. In the case of IoU=0.7, the results of Fcooper, DiscoNet, OPV2V, and V2X-ViT methods increased by 8.4\%, 4.5\%, 13.1\%, and 4.5\% respectively.

\begin{figure}[!t]
\includegraphics[width=3.5in]{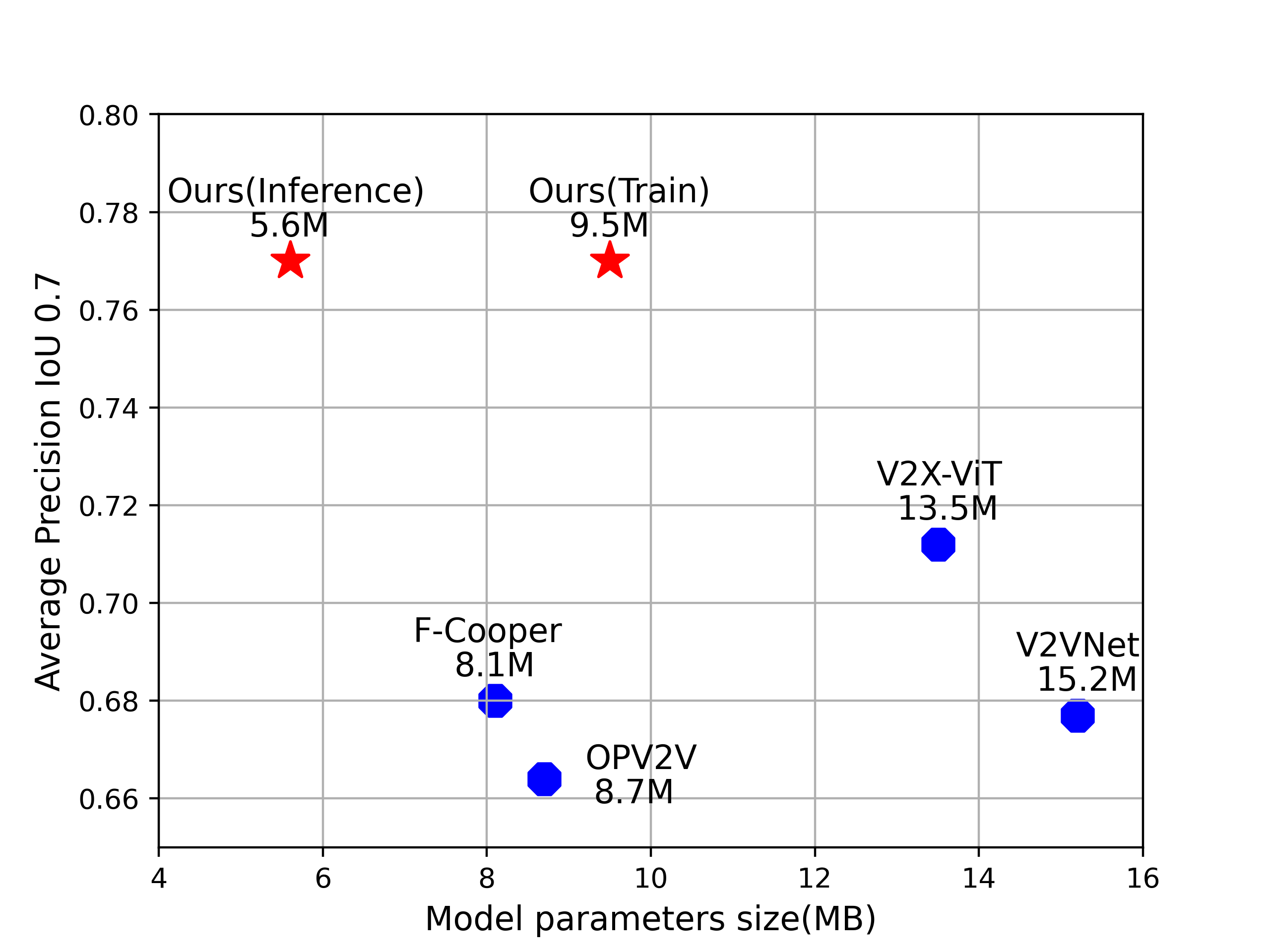}
\caption{Model parameter quantity and accuracy relationship graph.}
\label{fig_6}
\end{figure}

\begin{figure*}[!b]
\includegraphics[width=7in]{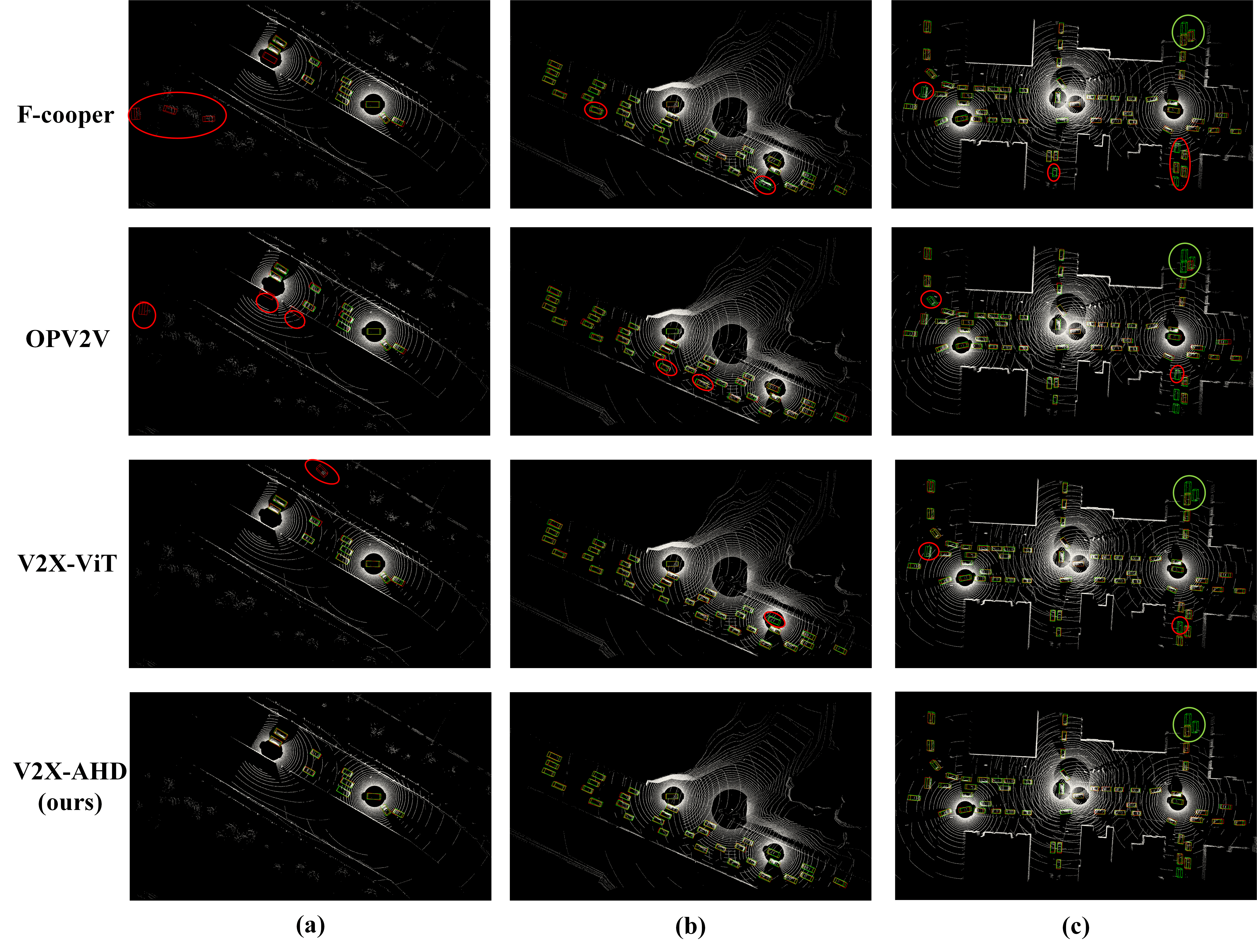}
\caption{Qualitative comparison in different conditions. Green and red bounding boxes represent ground truth and prediction, respectively. In addition, we use circles to illustrate the key of the figure.}
\label{fig_7}
\end{figure*}
Performance is determined jointly by the number of model parameters and the accuracy of prediction. Figure 6 illustrates the relationship between the parameter amounts of various algorithms and their precision. This article utilizes the knowledge distillation architecture, so there are two training and testing algorithm parameters. The figure demonstrates that this paper achieves the best possible result with the fewest number of test parameters. Combined with the quantitative analysis results, we discovered that the F-Cooper and OPV2V algorithms with small parameters significantly enhanced detection performance after the Spare Pillar module was replaced due to their simple fusion strategy. However, the V2X-ViT algorithm with many parameters has a slight improvement due to the increased complexity of the fusion strategy. The V2VNet with the greatest number of parameters even experienced negative growth. The preceding results demonstrate that the high-complexity fusion model will hinder the expression of features. The lightweight MSA feature fusion module presented in this paper is better suited for fusing single-view vehicle data features.

\subsubsection{Distillation temperature comparison}
The knowledge distillation structure employs soft training targets. Compared with manually labeled hard target labels, soft target labels contain more entropy, thereby increasing the difference between various features and allowing the student network to obtain more valuable information from the teacher network. Distillation temperature is a vital knowledge distillation hyperparameter, and Table 3 shows the AP results of the model at different distillation temperatures. When the temperature of distillation is low, detection precision increases as the temperature rises. The detection accuracy reached the highest value when the distillation temperature was ten, and then with the increase of distillation temperature, the detection accuracy showed a downward trend. According to the analysis, the inter-class differences of soft objects become smaller as distillation temperature increases, and the characteristics cannot be learned. Finally, we selected the optimum temperature as 10.

\begin{table}
\renewcommand\arraystretch{1.5}
\tabcolsep=0.5cm
\centering
\caption{Comparison of AP at different distillation temperatures.}
\label{tab3}
\begin{tabular}{ c c c } \hline
Temperature & AP(IoU=0.5)) & AP(IoU=0.7)\\ \hline
1 & 0.853 & 0.738\\
2 & 0.860 & 0.740\\
5 & 0.865 & 0.748\\
10 & \textbf{0.876} & \textbf{0.770}\\
15 & \textbf{0.876} & 0.763\\
20 & 0.877 & 0.759\\ \hline
\end{tabular}
\end{table}

\subsection{Qualitative analysis}
\subsubsection{Object detection result}
Figure 7 depicts the results of the detection process. We select three representative scenarios and compare the F-cooper, OPV2V, and V2X-viT algorithms' results. The green and red bounding boxes represent the ground truth and predicted results, respectively. Even when compared to the SOTA algorithm V2X-ViT, our algorithm has a higher degree of matching. The scene in Fig. 7(a) shows the false detection problem. All the algorithms recognize the vehicles in the scene. However, except for our algorithm, the other algorithms all identify some trees as vehicles, as shown in the red circle. It can be seen that the plug-in SparePillar has a more robust feature extraction performance than PointPillar. Figure 7 (b) depicts the occlusion detection results. The F-cooper algorithm and the OPV2V algorithm have a significant deviation of the detection bounding box angle or even occlusion-related missed detection. As shown by the red circle, the V2X-ViT algorithm has the issue of failed point cloud vehicle self-detection. However, the fused point cloud at this location has the shape of the vehicle. According to the analysis, the intermediate fusion algorithm cannot accurately synthesize the contradictory results. Additionally, it demonstrates that the complex fusion strategy will impede the correct expression of the extracted features. Figure 7(c) shows the scene with multiple intersections and agents. At the edge positions, such as the left and bottom sides of the scene, F-cooper, OPV2V, and V2X-ViT have undetected situations, as shown by the red circle marks. In areas where the point cloud is sparse, as indicated by the green circle, detection is also missed by all algorithms. Although there are actual bounding boxes, there are only a few point clouds, no effective vehicle features can be formed, and the failure to detect is a reasonable phenomenon.

\begin{figure*}[!t]
\includegraphics[width=7in]{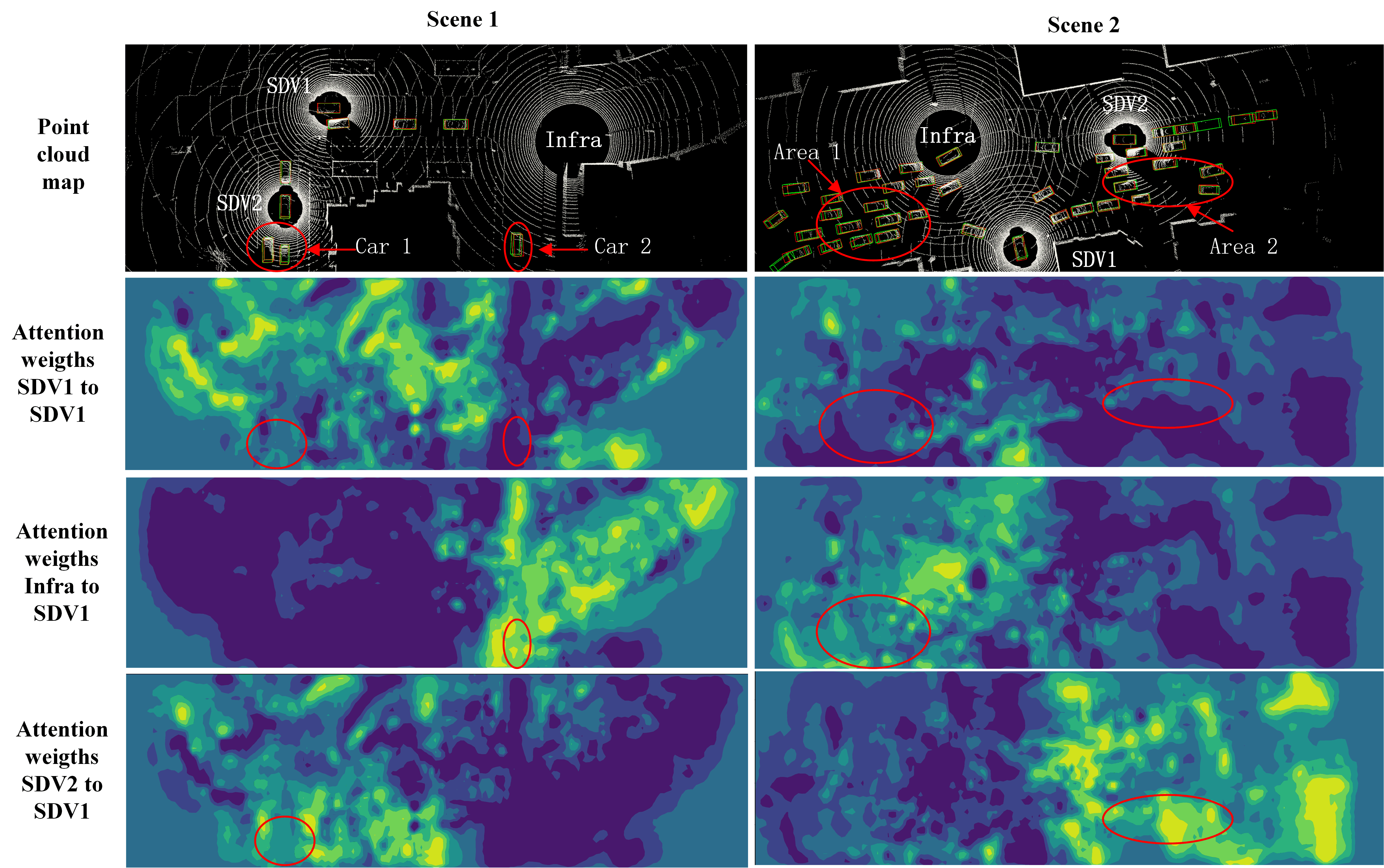}
\caption{Attention weight transfer. Scene1. Brighter parts of the image indicate higher attention to that area. Each Scene includes three agents: SDV1, SDV2, and, Infra, the SDV1 is the feature receiver.}
\label{fig_8}
\end{figure*}

\begin{table}[!t]
\renewcommand\arraystretch{1.8}
\tabcolsep=0.17cm

\caption{Ablation experimental results of different components in the v2xset dataset.}
\centering
\label{tab4}
\begin{tabular}{ c c c c c c } \hline
Baseline &  SparePillar & AHD & MSA & AP(IoU=0.5)) & AP(IoU=0.7)\\ \hline
$\checkmark$ & & & & 0.807 & 0.664\\
$\checkmark$ &$\checkmark$ & & & 0.864 & 0.754\\
$\checkmark$ & &$\checkmark$& & 0.858 & 0.736\\
$\checkmark$ &$\checkmark$ &$\checkmark$ & & 0.872 & 0.761\\
$\checkmark$ &$\checkmark$ &$\checkmark$ & $\checkmark$& \textbf{0.877} & \textbf{0.770}\\\hline
\end{tabular}
\end{table}
\subsubsection{Multi-view feature fusion}
Figure 8 shows the aggregation process between fused view data and visualizes the attention weights. Infra represents roadside infrastructure, while SDV1 and SDV2 represent self-driving vehicles. With SDV1 functioning as the data receiver, Infra and SVD2 transmit data to SDV1. The point cloud map and the attention map correspond; the brighter the color, the more attention the area has received. In Figure 8 (a), because SDV2 blocks Car1, SDV1 pays less attention to the position of Car1 and has a darker color, while the corresponding position of SDV2 has a darker color to the position of car1. However, car 2 is located at an intersection, the detection distance to SDV1 is relatively great, and there is occlusion. SDV1 pays little attention to it, whereas the roadside equipment at the corresponding location demonstrates high attention. In Figure 8 (b), SDV1 is blocked by the intersection, resulting in a restricted field of view. The highlighted area of the SDV1 attention map is limited to some intersections, while areas 1 and 2 are not focused due to occlusion. Infra and SDV2 in the area1 and area2 positions with higher attention. The attention-based fusion strategy MSA proposed in this paper can effectively synthesize the extracted point cloud features and ultimately achieve the objectives of extending the detection range of targets and controlling occlusion.

\subsection{Ablation experiments}
When IoU=0.7, the single module SparePillar, and AHD improve AP by 13.5\% and 10.8\%, respectively. The results indicate that SparePillar and the asymmetric distillation architecture can significantly improve perceptual accuracy. The MSA accuracy increased by 1.2\%, indicating that the fusion module is capable of enhancing the feature expression under higher perceptual accuracy than the self-attention mechanism. Under IoU=0.5 and IoU=0.7, the method presented in this paper is 8.7\% and 16.0\% more efficient than the baseline method. Since a single view has acquired multi-view features, it has achieved a greater improvement under the IoU=0.7 condition, which necessitates greater precision. The above ablation experimental results demonstrate the effectiveness of each component.

\section{Conclusion}
To solve the problem of low vehicle recognition accuracy caused by the unclear outline of the vehicle in the conventional single-vehicle point cloud target detection method. This paper proposes a multi-view vehicle-road collaborative perception framework V2X-AHD based on a distillation network. Compared with the conventional algorithm, the asymmetric heterogeneous distillation network AHD proposed in this paper can transfer the entire point cloud features of multiple views to a single view, improving object contour perception accuracy. The extractor Spare Pillar based on sparse convolution design improves the ability to extract features from point cloud data while simultaneously reducing the number of parameters. The feature fuser MSA performs feature layer fusion on multi-view feature data. This paper verifies the proposed method on the publicly available dataset V2XSet, achieving the best results, and demonstrates the effectiveness of the proposed method V2X-AHD through quantitative and qualitative analysis.

The proposed method has only been validated in a simulated environment, and the difference between vehicle-side and road-side data is negligible. In future work, we will collect real scene point cloud data, construct our own real scene data set, and examine the differences between vehicle and roadside point cloud data.

\vfill


\begin{thebibliography}{1}

\bibitem{1}
H.Wang, Y Chen, Y Cai, et al. "SFNet-N: An improved SFNet algorithm for semantic segmentation of low-light autonomous driving road scenes." {\it{IEEE Transactions on Intelligent Transportation Systems,}} vol. 23, no. 11, pp.21405-21417, 2022.

\bibitem{2}
Y. Cai, T. Luan, H. Gao, H. Wang, L. Chen, Y. Li, M. A. Sotelo, and Z. Li, "YOLOv4-5D: An effective and efficient object detector for autonomous driving."{\it{ IEEE Transactions on Instrumentation and Measurement,}} vol. 70, pp. 1-13, 2021. 

\bibitem{40}
A. Lombard, A. Noubli, A. Abbas-Turki, et al. "Deep Reinforcement Learning Approach for V2X Managed Intersections of Connected Vehicles". {\it{IEEE Transactions on Intelligent Transportation Systems,}} 2023.
\bibitem{41}
X. Chen , S. Leng, J. He, et al."The Upper Bounds of Cellular Vehicle-to-Vehicle Communication Latency for Platoon-Based Autonomous Driving". {\it{IEEE Transactions on Intelligent Transportation Systems,}} 2023.
\bibitem{42}
L. Zhao, Y. Song, C. Zhang, et al. "T-gcn: A temporal graph convolutional network for traffic prediction." {\it{IEEE Transactions on Intelligent Transportation Systems,}} vol.21, no.9, pp.3848-3858, 2019.
\bibitem{43}
W. Chen ,J. Zhao, W. L. Zhao, et al. "Shape-Aware Monocular 3D Object Detection." {\it{IEEE Transactions on Intelligent Transportation Systems,}} 2023.
\bibitem{44} 
Y. Kim, S. Kim, S. Sim, et al. "Boosting Monocular 3D Object Detection With Object-Centric Auxiliary Depth Supervision." {\it{IEEE Transactions on Intelligent Transportation Systems,}} vol.24, no. 2, pp. 1801-1813, 2022.

\bibitem{14}
C. He, H. Zeng, J. Huang, et al. "Structure aware single-stage 3d object detection from point cloud." {\it{Proceedings of the IEEE/CVF conference on computer vision and pattern recognition,}} pp. 11873-11882, 2020.
\bibitem{16}
Y. Yan, Y. Mao, B. Li.  "Second: Sparsely embedded convolutional detection." {\it{Sensors,}} vol. 18, no. 10, pp. 3337, 2018.

\bibitem{17}
W. Zheng, W. Tang, S. Chen, et al. "Cia-ssd: Confident iou-aware single-stage object detector from point cloud." {\it{Proceedings of the AAAI conference on artificial intelligence,}} vol. 35, no. 4, pp. 3555-3562, 2021.
\bibitem{20}
H. Kuang, B. Wang, J. An, et al. "Voxel-FPN: Multi-scale voxel feature aggregation for 3D object detection from LIDAR point clouds."{\it{ Sensors,}} vol. 20, no.3, pp. 704,2020.
\bibitem{23}
C. R. Qi, H. Su, K. Mo, et al. "Pointnet: Deep learning on point sets for 3d classification and segmentation." {\it{Proceedings of the IEEE conference on computer vision and pattern recognition,}} pp. 652-660,2017.

\bibitem{25}
Q. Chen, S, Tang, Q, Yang, et al. "Cooper: Cooperative perception for connected autonomous vehicles based on 3d point clouds." {\it{2019 IEEE 39th International Conference on Distributed Computing Systems (ICDCS). IEEE,}}pp. 514-524,2019.
\bibitem{8}
R. Xu, H, Xiang, Z. Tu, et al. "V2x-vit: Vehicle-to-everything cooperative perception with vision transformer."{\it{  European conference on computer vision,}} pp.107-124,2022.


\bibitem{26}
Q. Chen, X. Ma, S. Tang, et al. "F-cooper: Feature based cooperative perception for autonomous vehicle edge computing system using 3D point clouds." {\it{Proceedings of the 4th ACM/IEEE Symposium on Edge Computing,}} pp. 88-100, 2019.

\bibitem{27}
T. H. Wang, S. Manivasagam, M. Liang, et al. "V2vnet: Vehicle-to-vehicle communication for joint perception and prediction." {\it{Computer Vision–ECCV 2020: 16th European Conference, Glasgow, UK,}}pp. 605-621, 2020.

\bibitem{28}
R. Xu, H. Xiang, X. Xia, et al. "Opv2v: An open benchmark dataset and fusion pipeline for perception with vehicle-to-vehicle communication." {\it{2022 International Conference on Robotics and Automation (ICRA). IEEE,}}pp. 2583-2589, 2022.

\bibitem{29}
A. Rauch, F. Klanner, R. Rasshofer, et al. "Car2x-based perception in a high-level fusion architecture for cooperative perception systems." {\it{2012 IEEE Intelligent Vehicles Symposium. IEEE,}} pp. 270-275, 2012.

\bibitem{30}
Z. Y. Rawashdeh, Z. Wang. "Collaborative automated driving: A machine learning-based method to enhance the accuracy of shared information." {\it{2018 21st International Conference on Intelligent Transportation Systems (ITSC). IEEE,}} pp. 3961-3966, 2018.
\bibitem{39}
T. Luo, L. Chen, T. Luan, et al. "A Vehicle-to-Infrastructure beyond Visual Range Cooperative Perception Method Based on Heterogeneous Sensors." {\it{Energies,}}  vol.15 no.21 pp: 7956, 2022.



\bibitem{3}
R. Anil,G. Pereyra, A. Passos, et al. "Large scale distributed neural network training through online distillation." {\it{ arXiv preprint,}} arXiv:1804.03235, 2018. 

\bibitem{4}
G. Hinton, O. Vinyals, J. Dean." Distilling the knowledge in a neural network."{\it{  arXiv preprint,}} arXiv:1503.02531,2015.

\bibitem{5}
A. Romero, N. Ballas,S. Kahou, et al. "Fitnets: Hints for thin deep nets." {\it{ arXiv preprint,}} arXiv:1412.6550, 2014.

\bibitem{6}
G. Shi, R. Li, C. Ma. "PillarNet: High-Performance Pillar-based 3D Object Detection." {\it{ arXiv preprint,}} arXiv:2205.07403, 2022.

\bibitem{7}
B. Graham, V.D. Maaten, L.: "Submanifold sparse convolutional networks".{\it{  arXiv preprint,}} arXiv:1706.01307,2017.



\bibitem{10}
Y. Liu, K. Chen, C. Liu, et al. "Structured knowledge distillation for semantic segmentation." {\it{Proceedings of the IEEE/CVF conference on computer vision and pattern recognition,}} pp. 2604-2613, 2019.

\bibitem{11}
Y. Wang, A. Fathi, J. Wu, et al. "Multi-frame to single-frame: Knowledge distillation for 3d object detection." {\it{arXiv preprint,}} arXiv:2009.11859, 2020. 

\bibitem{12}
X. Jin, C. Lan, W. Zeng, et al. "Uncertainty-aware multi-shot knowledge distillation for image-based object re-identification." {\it{Proceedings of the AAAI Conference on Artificial Intelligence,}} vol. 34 no. 07, pp. 11165-11172, 2020.

\bibitem{13}
Y. Li, S. Ren, P. Wu, et al. "Learning distilled collaboration graph for multi-agent perception." {\it{Advances in Neural Information Processing Systems,}} vol. 34, pp. 29541-29552,2021.


\bibitem{15}
Y. Hu, Z. Ding, R. Ge, W. Shao , L. Huang, K. Li, Q. Liu  "Afdetv2: Rethinking the necessity of the second stage for object detection from point clouds." {\it{In Proceedings of the AAAI Conference on Artificial Intelligence,}} vol. 36, no. 1, pp. 969-979, 2022.



\bibitem{18}
Y. Zhou, O. Tuzel. "Voxelnet: End-to-end learning for point cloud based 3d object detection." {\it{Proceedings of the IEEE conference on computer vision and pattern recognition,}} pp. 4490-4499, 2018.

\bibitem{19}
A. H. Lang, S. Vora, H. Caesar, et al. "Pointpillars: Fast encoders for object detection from point clouds." {\it{Proceedings of the IEEE/CVF conference on computer vision and pattern recognition,}} pp. 12697-12705, 2019.


\bibitem{21}
M. Ye, S. Xu, T. Cao. "Hvnet: Hybrid voxel network for lidar based 3d object detection." {\it{Proceedings of the IEEE/CVF conference on computer vision and pattern recognition,}} pp. 1631-1640, 2020.

\bibitem{22}
J. Noh, S. Lee, B. Ham. "Hvpr: Hybrid voxel-point representation for single-stage 3d object detection." {\it{Proceedings of the IEEE/CVF conference on computer vision and pattern recognition,}} pp.14605-14614, 2021.





\bibitem{31}
I. Goodfellow, D. Warde-Farley, M. Mirza, et al. "Maxout networks." {\it{International conference on machine learning. PMLR,}} pp. 1319-1327, 2013.

\bibitem{32}
A. Vaswani, N. Shazeer, N. Parmar, et al. "Attention is all you need." {\it{Advances in neural information processing systems,}} vol 30, 2017, 30.

\bibitem{33}
M. Schlichtkrull, T. N. Kipf, P. Bloem, et al. "Modeling relational data with graph convolutional networks." {\it{The Semantic Web: 15th International Conference, ESWC 2018,}} pp. 593-607, 2018.

\bibitem{34}
K. Simonyan, A. Zisserman . "Very deep convolutional networks for large-scale image recognition." {\it{arXiv preprint,}} arXiv:1409.1556, 2014.

\bibitem{35}
A. Dosovitskiy, G. Ros, F. Codevilla, et al. "CARLA: An open urban driving simulator." {\it{Conference on robot learning. PMLR,}} pp 1-16, 2017.

\bibitem{36}
R.Xu, Y. Guo, X. Han, et al. "OpenCDA: an open cooperative driving automation framework integrated with co-simulation." {\it{2021 IEEE International Intelligent Transportation Systems Conference (ITSC). IEEE,}}pp. 1155-1162, 2021.

\bibitem{37}
D. P. Kingma, J.Ba. "Adam: A method for stochastic optimization." {\it{arXiv preprint,}} arXiv:1412.6980, 2014.

\bibitem{38}
T. Y. Lin, P. Goyal, R.Girshick, et al. "Focal loss for dense object detection." {\it{Proceedings of the IEEE international conference on computer vision,}}pp. 2980-2988, 2017.







\end{thebibliography}
\end{document}